\documentclass{IOS-Book-Article}
\usepackage{mathptmx}
\usepackage{url}
\usepackage{booktabs}
\usepackage{tabularx}
\usepackage{multirow}
\usepackage{amsmath}
\usepackage{graphicx}
\usepackage{subcaption}
\usepackage{siunitx}
\newcommand\blfootnote[1]{%
  \begingroup
  \renewcommand\thefootnote{}\footnote{#1}%
  \addtocounter{footnote}{-1}%
  \endgroup
}

\begin{document}
\begin{frontmatter}              

\title{Label Indeterminacy in AI \& Law}
\runningtitle{IOS Press Style Sample}

\author[A]{\fnms{Cor} \snm{Steging}}
\author[B]{\fnms{Tadeusz} \snm{Zbiegień}}

\runningauthor{Steging \&  Zbiegień.}
\address[A]{Bernoulli Institute of Mathematics, Computer Science and Artificial Intelligence, University of Groningen}
\address[B]{Department of Legal Theory, Jagiellonian University}

\begin{abstract}
Machine learning is increasingly used in the legal domain, where it typically operates retrospectively by treating past case outcomes as ground truth. However, legal outcomes are often shaped by human interventions that are not captured in most machine learning approaches. A final decision may result from a settlement, an appeal, or other procedural actions. This creates label indeterminacy: the outcome could have been different if the intervention had or had not taken place. We argue that legal machine learning applications need to account for label indeterminacy. Methods exist that can impute these indeterminate labels, but they are all grounded in unverifiable assumptions. In the context of classifying cases from the European Court of Human Rights, we show that the way that labels are constructed during training can significantly affect model behaviour. We therefore position label indeterminacy as a relevant concern in AI \& Law and demonstrate how it can shape model behaviour. 
\end{abstract}

\end{frontmatter}

\thispagestyle{empty}
\pagestyle{empty}
\blfootnote{This manuscript has been accepted for presentation as a short paper at the 38th International Conference on Legal Knowledge and Information Systems (JURIX) in Turin, December 9 to 11 of 2025.}

\section{Introduction}
With recent advances in machine learning, artificial intelligence is increasingly deployed in law, yet many approaches inherently contain characteristics that are incompatible with legal tasks. Examples include the intrinsic retrospective nature of machine learning, and the inability to faithfully explain their reasoning~\cite{bench2020need, StegingJURIX2021}, and generative AI models suffer from issues such as hallucinations~\cite{Dahl2024LargeLegalFictions}.

One crucial machine learning issue that has not yet been explored in the legal domain is \textit{label indeterminacy}~\cite{Schoeffer2025Perils}, where the labels of certain training instances are inherently unknown because an intervention by a decision maker has affected the final outcome. An example comes from medicine, where doctors must decide whether to continue or withhold life-sustaining treatment. If treatment is withheld, the patient does not survive, and it is impossible to know whether they would have survived had treatment continued. The resulting data is therefore \textit{selectively labelled}, as the observed outcomes are contingent on the decisions of the doctor~\cite{Lakkaraju2017Selective}. Historical cases where treatment was withheld thus yield indeterminate labels, and any machine learning model trained on such data must take this indeterminacy into account. Research from other domains has shown that the way that we treat these indeterminate labels can have drastic effects on the behaviour and outcomes of AI models~\cite{Schoeffer2025Perils}. 

The legal domain also knows many scenarios in which the outcome has been affected by an intervention, and thus where label indeterminacy is at play. A famous example is that of granting bail: if an individual is not granted bail, we do not know how they would have acted if they were given bail~\cite{Lakkaraju2017Selective}. If one were to train AI models on the historical cases, one would need to account for the pretrial detention cases, as those labels are indeterminate. If this is not accounted for, the model effectively learns to replicate past detention decisions rather than predicting the defendant’s likelihood of returning for a court appearance, thereby potentially perpetuating historical bias, as illustrated by concerns raised as in the COMPAS system~\cite{COMPAS}. Excluding all non-bail cases is likewise problematic, since they do not constitute a random sample of bail cases.

Indeterminate labels can also arise in the case of settlements, where we do not know what the litigation outcome would have been if the case had been taken to court. Excluding settled cases from the training data can yield an unrepresentative model, as cases are not settled at random, and thus litigated cases alone may not be an accurate representative sample of all cases.
Taking this one step further, in legal judgment prediction where cases could have been appealed but the judgment was accepted, we cannot know what the outcome would have been if the case had been appealed. The original judgment then may not reflect what an appellate court would have decided. Yet excluding original judgments from training data is problematic, since not all cases are or can be appealed based on the content.

In this paper, we critically examine the legal task of court case predictions. 
We examine when and why decisions in non-appealed cases should be considered indeterminate rather than ground truth, and what methods exist that can be used to impute the labels of these indeterminate cases.
In an experiment using court case data from the European Court of Human Rights (ECtHR), we show that the commonly used methods to handle label indeterminacy can lead to different prediction behaviours of the models that we train.

\section{Background}
Machine learning approaches to automated legal prediction constitute an established subfield, with various models attempting to classify case outcomes across different legal domains~\cite{Katz2017SupremeCourtPrediction, Aletras2016ECHRPrediction, chalkidis-etal-2019-neural, Medvedeva2020ECHRPrediction,cui2022surveylegaljudgmentprediction}. Most of the previous research has used past case outcomes as fixed objective labels. This relies on an often implicit and rarely discussed modelling assumption that the final judgment represents an unambiguous resolution of the dispute and an objectively determined outcome. 
This assumption is particularly problematic in the context of label indeterminacy, where legal outcomes may depend on interventions such as appeals, settlements, or procedural decisions \cite{4a5c48d0-8e3c-3eeb-9ba2-36027f49b6d0,d95bc569-4324-3be1-91ed-971e8c56f0f8}.
As the sub-field of automated legal prediction matures, core theoretical questions about what is meant by `legal prediction' have emerged, along with considerations that must be addressed in practical applications~\cite{medvedeva2022rethinking, steging2023taking}.  
In this context, the assumption that past case outcomes are always fixed and objective warrants revisiting, both from the computational and jurisprudential perspectives.

To clarify the scope of this paper, it is first useful to distinguish label indeterminacy from \textit{selection bias}, in which datasets capture only a non-representative slice of all disputes~\cite{medvedeva-mcbride-2023-legal}. 
Both phenomena affect legal prediction, but in different ways.
For example, excluding settled cases from a dataset introduces selection bias, as the remaining cases may not be representative of all disputes. 
In contrast, label indeterminacy arises when the outcome itself is unknowable: when settled cases are included in the dataset, the true court outcome of those cases cannot be determined.
Consequently, a `final judgment' in a dataset may therefore not reflect the whole story, depending on the qualities of a particular process. 

Distinct from these data-centric issues is the discussion on \textit{legal indeterminacy}, a jurisprudential debate that should not be conflated with label indeterminacy.   The indeterminacy debate developed from the realist critique of legal formalism~\cite{Hasnas1995, Spaic2023Formalism}. It argues that legal reasoning often fails to yield a single inevitable answer because legal rules and principles are underdetermined \cite{Singer1984, 3fe54c1e-8e72-3380-95cd-4efbf3b01459} . Critical legal studies scholars have further emphasized that the law functions as a political practice with outcomes shaped by existing power dynamics rather than by mechanistic rule-following \cite{3fe54c1e-8e72-3380-95cd-4efbf3b01459, f6fb234a-5b25-320b-8f5a-0d03bd460ff2}. 

One of the most influential rebuttals to strong forms of legal indeterminacy comes from Dworkin~\cite{dworkin1977}, who argued that even in hard cases there is, in principle, a `right answer' discoverable through proper interpretation (see also~\cite{whitman2008}). This debate was of historical significance for the development of modern thinking about law and continues to influence it today.

More broadly, and apart from the debate on legal indeterminacy, scholars have noted that the notion of `truth’ in law is often constituted within institutional and argumentative practices. For example, in the legal context Patterson challenges the correspondence view of truth~\cite{Aquinas2007, coleman1995truth}, and instead, locates truth in the accepted forms of legal argument. This makes `truth' a matter of institutional endorsement rather than strict fact~\cite{Patterson1996-PATLAT-4}. This linguistic-practice approach emphasizes that what counts as `true' in law can be contingent on shared argumentative norms, rather than on an external, fixed ground truth ~\cite{Patterson1996-PATLAT-4}. The argumentative perspective has also gained traction within the AI \& Law domain~\cite{Governatori2022ThirtyYears, Sartor2022ThirtyYears,Villata2022ThirtyYears} .

While our aim is not to settle theoretical disputes about the indeterminacy of law or define what the ground truth in the legal domain actually means, there are grounds to argue that outcomes of legal disputes are not purely mechanical outputs of legal rules but are contingent and, at times, reversible products of procedural frameworks, substantive norms, and other difficult-to-capture contextual factors.

Treating all labels as static ground truths therefore risks misrepresenting the very phenomena machine learning models seek to capture.
In our upcoming experiments, we show that the way we treat indeterminate labels can lead to major differences in model behaviour. Moreover, it has the potential to reinforce historical biases and ultimately undermine the reliability of AI applications in law.

\section{European Court of Human Rights}
We investigate label indeterminacy from a legal perspective in the domain of the European Court of Human Rights (ECtHR), where we develop a machine learning model to predict case outcomes. Predicting ECtHR case outcomes is a challenging and well-studied academic task~\cite{medvedeva2022rethinking}, making it a suitable context for our study.

The ECtHR has several judicial formations, with most cases initially heard by a Chamber, a seven-judge panel that decides cases by majority vote~\cite{echr_rules}. In certain circumstances, either by referral or after relinquishment of jurisdiction by a Chamber, a case can be sent to the Grand Chamber, a 17-judge body that serves as the Court’s final authority. Referral is not automatic, and many Chamber judgments remain final without Grand Chamber review. This procedural structure raises a key question for modeling: if a case is not referred, should the Chamber judgment be treated as definitive? Legally, the answer is nuanced: Chamber decisions are binding unless overturned or amended, yet they lack the authority of Grand Chamber judgments. From a machine learning perspective, we cannot model both simultaneously, as their decisions can conflict, for example, when a case is appealed and receives a different ruling from the Grand Chamber. 

In this study, we decide to model the decisions of the Grand Chamber, the Court’s highest authority, and treat these decisions as the ground truth. Consequently, Chamber decisions are considered indeterminate, as they do not necessarily align with the Grand Chamber’s rulings. This framework allows us to investigate how different strategies for handling label indeterminacy, such as treating Chamber decisions as provisional or excluding them entirely, affect classifier behaviour. 

We focus on cases brought under Article 6 of the European Convention on Human Rights, which concerns the right to a fair trial and constitutes by far the largest category of applications to the Court. We frame the task as a binary classification problem: determining whether a violation of Article 6 occurred (yes or no) based on the facts as documented by the ECtHR post-ruling. Technically, this falls into the larger category of outcome categorization tasks in legal NLP~\cite{medvedeva2022rethinking}.

\section{Method}
This section details the setup of our experiment, including the dataset, the models employed, and the various existing strategies for handling the indeterminate labels of the Chamber cases.

\subsection{Dataset}
The dataset for our study was obtained from the ECHR Open Data project~\cite{ECHROD} and comprises ECtHR cases from 1968 to 2023. 
It includes 5,902 Chamber cases and 191 Grand Chamber cases related to Article 6. We format the dataset such that each case contains the facts section in natural language, the outcome label (violation or non-violation), and an indication of whether it was judged by the Chamber or the Grand Chamber. Our model is trained on the text of the facts section to predict case outcomes. To account for temporal effects in the law~\cite{steging2023taking}, we split the dataset by year of judgment: cases before 2015 form the training set, while cases from 2015 onward constitute the test set. The resulting dataset distributions are summarized in Table~\ref{tbl:split_summary}. 

\begin{table}[ht]
\centering
\caption{Dataset distribution, where cases from 2015 and later are used as test set. Seven balanced train sets are generated from the train set.}
\renewcommand{\arraystretch}{1.3}
\begin{tabular}{l c c c c c c c}
\toprule
\multirow{2}{*}{\textbf{Dataset}} & \multirow{2}{*}{\textbf{Total}} &
\multicolumn{2}{c}{\textbf{Test Set}} &
\multicolumn{2}{c}{\textbf{Train Set}} &
\multicolumn{2}{c}{\textbf{Balanced Train Sets}} \\
\cmidrule(lr){3-4} \cmidrule(lr){5-6} \cmidrule(lr){7-8}
& & \textbf{Size} & \textbf{\% Violation} & \textbf{Size} & \textbf{\% Violation} & \textbf{Size} & \textbf{\% Violation} \\
\midrule
Grand Chamber & 191   & 34  & 47.06\% & 157  & 63.06\% & 116  & 50.00\% \\
Chamber       & 5902  & 871 & 69.80\% & 5031 & 87.48\% & 1376 & 50.00\% \\
\bottomrule
\end{tabular}
\label{tbl:split_summary}
\end{table}

\subsection{Model and preprocessing}
Because ECtHR case files are long, we use an architecture capable of handling large documents. We employ a Longformer~\footnote{\url{https://huggingface.co/allenai/longformer-base-4096}}, which supports inputs of up to 4,096 tokens~\cite{Beltagy2020Longformer}. Analysis of our training data shows that Chamber cases average approximately 2,315 tokens, while Grand Chamber cases average 5,638 tokens (Table~\ref{tbl:token_stats}). Only 85.3\% of Chamber cases and 45.9\% of Grand Chamber cases fit within the 4,096-token limit. We first apply light preprocessing to remove structural noise and unnecessary tokens, followed by head-tail truncation for longer cases to make them fit within the token limit (Table~\ref{tbl:token_stats}).

\begin{table}[ht]
\centering
\caption{Token Statistics for Chamber and Grand Chamber Cases}
\begin{tabular}{llccccc}
\toprule
\textbf{Dataset} & \textbf{Metric} & \textbf{Cases} & \textbf{Max} & \textbf{Mean} & \textbf{Median} & \textbf{\% $\leq$ 4096} \\
\midrule
\multirow{2}{*}{Chamber} 
& Raw & 5031 & 75456 & 2317.9 & 1303.0 & 85.3 \\
& Preprocessed & 5031 & 4096 & 1756.0 & 1303.0 & 100 \\
\midrule
\multirow{2}{*}{Grand Chamber} 
& Raw & 157 & 22029 & 5638.6 & 4466.0 & 45.9 \\
& Preprocessed & 157 & 4096 & 3275.6 & 4096.0 & 100 \\
\midrule
\multirow{2}{*}{\textbf{Total}} 
& Raw & 5188 & 75456 & 2442.5 & 1354.0 & 84.1 \\
& Preprocessed & 5188 & 4096 & 1802.0 & 1353.0 & 100 \\
\bottomrule
\end{tabular}
\label{tbl:token_stats}
\end{table}

As shown in Table~\ref{tbl:split_summary}, the label distribution is heavily skewed toward violations. To mitigate bias toward the majority class, we balance the datasets to a 50\%-50\% distribution. For both Chamber and Grand Chamber cases, we create seven balanced training datasets. Each dataset includes all non-violation cases and an equal number of violation cases, with minimal overlap in violation cases across datasets. This results in seven balanced Chamber training datasets containing 1,376 cases each, and seven balanced Grand Chamber datasets containing 116 cases each (Table~\ref{tbl:split_summary}). We retain the natural imbalance of the test set, accounting for it appropriately during evaluation.

\subsection{Imputing indeterminate labels}
In our experiment, we model Grand Chamber decisions and therefore treat Chamber decisions as indeterminate, since the outcomes could have differed if these cases had been appealed to the Grand Chamber. We apply and compare nine methods for imputing these indeterminate labels to investigate their effects on model behaviour. The label imputation methods largely follow previous work by Schoeffer et al.~\cite{Schoeffer2025Perils}, and an overview is provided in Table~\ref{tbl:methods}. For each method, we briefly describe its procedure and the unverifiable assumptions underlying its application.

\begin{table}[ht]
\centering
\scriptsize
\setlength{\tabcolsep}{6pt} 
\renewcommand{\arraystretch}{1.1} 
\begin{tabular}{p{0.09\textwidth} p{0.16\textwidth} p{0.29\textwidth} p{0.31\textwidth}}
\toprule
\textbf{ID} & \textbf{Name} & \textbf{Description} & \textbf{Assumption / Risk} \\
\midrule
\texttt{$corr$} & Correct Chamber & Includes Chamber cases without altering their labels & Assumes past Chamber decisions reflect Grand Chamber decisions \\
\addlinespace[1mm]
\texttt{$obs$} & Observed only & Trains only on Grand Chamber cases & Assumes missing at random; fails if appeal is outcome-related \\
\addlinespace[1mm]
\texttt{$obs+ip$} & Observed + IP & Trains only on Grand Chamber cases and applies inverse propensity weights to correct for sample bias & Assumes positivity and no unobserved confounders \\
\addlinespace[1mm]
\texttt{$nn$} & Nearest neighbor & Imputes Chamber labels using the most similar Grand Chamber case & Assumes a valid similarity metric \\
\addlinespace[1mm]
\texttt{$exp_{all}$} & All experts & Uses each expert label individually with equal weighting & Assumes all expert inputs are equally valid, including conflicting information \\
\addlinespace[1mm]
\texttt{$exp_{avg}$} & Average expert & Averages multiple expert assessments into one label & Assumes the mean is meaningful; removes disagreement information \\
\addlinespace[1mm]
\texttt{$exp_{max}$} & Max expert & Labels violation if at least one judge voted for it & Assumes a judge voting for violation is correct; dissenting votes disregarded \\
\addlinespace[1mm]
\texttt{$exp_{min}$} & Min expert & Labels non-violation if at least one judge voted against it & Assumes a judge voting against violation is correct; other votes disregarded \\
\addlinespace[1mm]
\texttt{$exp_{agr}$} & Expert agree & Keeps only cases with full expert consensus & Assumes unanimous agreement implies correctness; ambiguous cases excluded \\
\bottomrule
\end{tabular}
\caption{Overview of methods used to impute the labels of Chamber cases.}
\label{tbl:methods}
\end{table}

\textit{Correct Chamber ($corr$)}. In this approach, the default in the legal judgment prediction literature, the labels of Chamber cases are left unchanged, under the assumption that they reflect the decisions the Grand Chamber would have made. However, this assumption is not always valid, as the Grand Chamber may reach different conclusions than the Chamber, as evident in cases that were appealed and subsequently overturned. \\

\textit{Observed only ($obs$)}. Here, all Chamber cases are excluded from the training dataset, leaving only Grand Chamber cases as ground truth. This method implicitly assumes that Chamber cases are missing at random. However, because Chamber cases are not a random subset and appeals may be outcome-dependent, excluding them can introduce bias into the training data.

\textit{Observed only with inverse propensity weighting ($obs+ip$)}. In this method, chamber cases are excluded, but inverse propensity weighting is applied to Grand Chamber cases during training to reduce sampling bias. As a result, Grand Chamber cases that are similar to Chamber cases are assigned higher weights. This method relies on two key assumptions. First, that every case has a non-zero probability of being appealed to the Grand Chamber (positivity), which does not hold in practice. Second, that there are no unobserved confounders, meaning the decision to appeal depends only on observable factors in the dataset, an assumption that cannot be verified.

\textit{Nearest neighbor ($nn$)}. In this method, we do include Chamber cases, and their labels are imputed based on similar Grand Chamber cases. Each case is first converted to embeddings using the Longformer. The embeddings are normalized, and the similarity between each Chamber case and all Grand Chamber cases is computed using the dot product. Each Chamber case is then assigned the label of the Grand Chamber case with the highest similarity. This method assumes that the similarity metric captures meaningful legal similarity and that cases deemed similar according to this metric would result in the same decision.

\textit{All experts ($exp_{all}$)}. This and the following approaches require expert labels for Chamber cases. Each case is duplicated $n$ times, once for each expert, and assigned that expert’s label. Each instance receives a sample weight of $\frac{1}{n}$. In our setting, the seven judges on a Chamber case are treated as experts, with their votes for or against a violation serving as labels weighted by $\frac{1}{7}$. This method assumes all expert assessments are equally valid, even if they conflict, so the model is trained on potentially inconsistent information.

\textit{Average experts ($exp_{avg}$)}. Labels are represented as continuous values, calculated as the number of judges voting for a violation divided by the total number of judges $n$ (seven in our case). A limitation of this approach is that it obscures individual disagreements, and the notion of an `average expert' may lack validity in real-world applications.

\textit{Max experts ($exp_{max}$)}. A case is labeled as a violation if at least one judge votes for violation, and as non-violation otherwise. This corresponds to an `optimistic' perspective in the literature, assuming that any indication supporting a positive label is correct. It presumes that judges voting for violation are correct and that dissenting judges are incorrect.

\textit{Min experts ($exp_{min}$)}. This method adopts a `pessimistic' perspective, labeling a case as non-violation if at least one judge votes against violation. It assumes that judges voting against violation are correct while the others are not.

\textit{Experts agree ($exp_{agr}$)}. Only Chamber cases with unanimous votes, either for or against violation, are retained. The underlying assumption is that unanimous decisions are correct, which may not always hold in practice.


\subsection{Experimental set-up}
We apply each of the nine label imputation methods to each of our seven balanced training sets, resulting in a total of 63 distinct datasets for training the Longformer model. Following light empirical parameter tuning, all models are trained with a batch size of 2 and a learning rate of $2 \times 10^{-5}$ for 3 epochs. Further details regarding model implementation and training procedures are available in our GitHub repository.~\footnote{\url{https://github.com/CorSteging/Label-Indeterminacy-in-AI-Law}}

\section{Results}

\begin{figure}[ht]
    \centering
    \includegraphics[width=\linewidth]{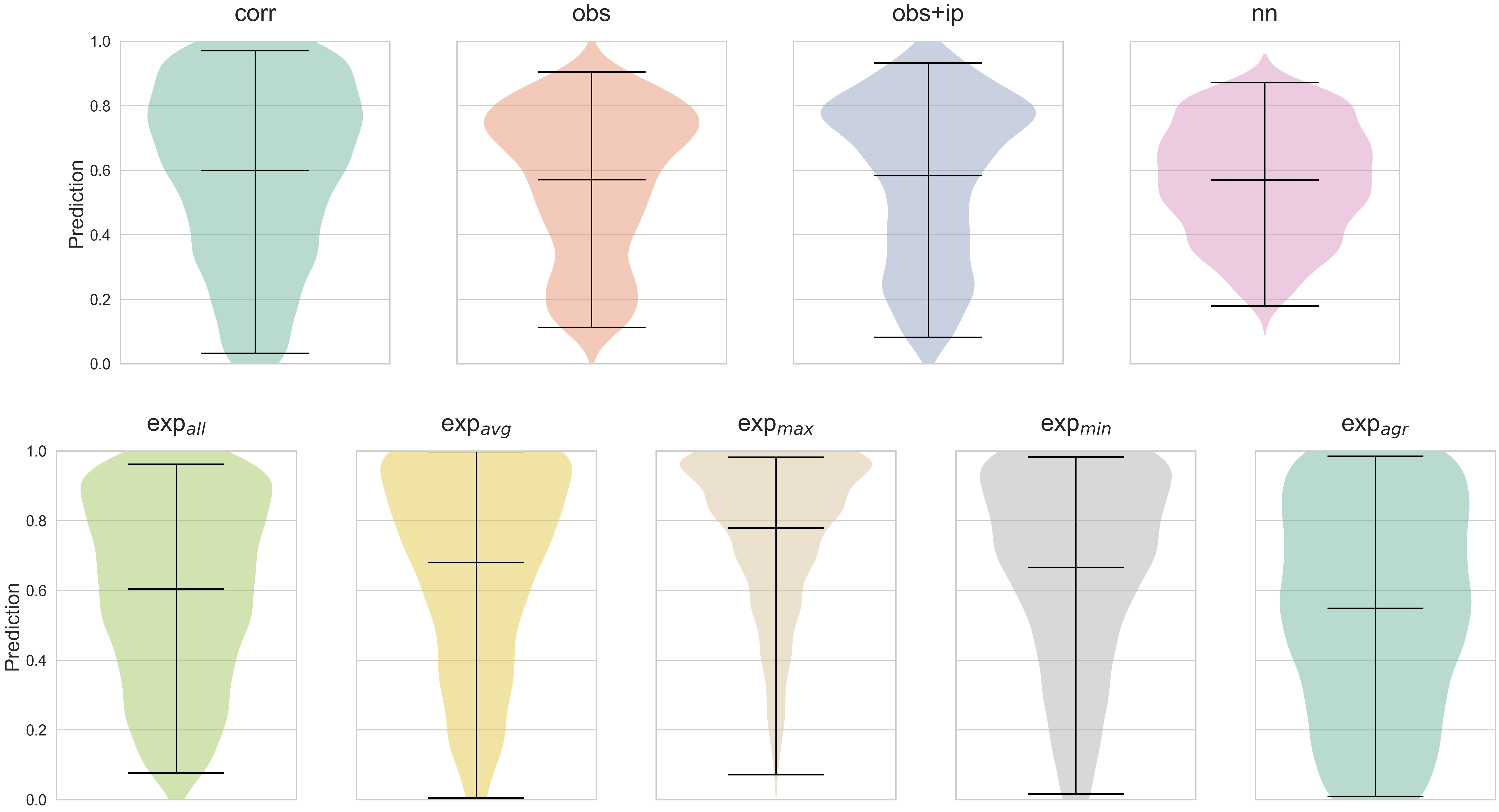}
    \caption{Violin plots displaying the distribution of the predictions for models trained using different label imputation methods.}
    \label{fig:violin}
\end{figure}

\begin{table}[ht]
\small
\setlength{\tabcolsep}{4pt}
\centering
\caption{Matthews Correlation Coefficient (MCC) by Method for Grand Chamber cases (ground truth labels) and Chamber cases (indeterminate labels)}
\label{tbl:mcc_combined}
\begin{tabular}{rccccccccc}
\toprule
& \multicolumn{9}{c}{\textbf{Grand chamber test set}} \\
\cmidrule(lr){2-10}
 & corr & obs & obs+ip & nn & expall & expavg & expmax & expmin & expagr \\
\midrule
Mean MCC  & -4.67 & 18.74 & 23.11 & 5.49 & -5.86 & 1.89 & 1.42 & 3.86 & 1.29 \\
Std Dev   &  6.10 & 19.96 & 17.36 & 10.97 & 14.71 & 17.31 & 19.20 & 16.55 & 13.40 \\
\midrule
& \multicolumn{9}{c}{\textbf{Chamber test set}} \\
\cmidrule(lr){2-10}
 & corr & obs & obs+ip & nn & expall & expavg & expmax & expmin & expagr \\
\midrule
Mean MCC  & 14.43 & 0.14 & 1.23 & 9.10 & 16.17 & 14.47 & 11.58 & 14.37 & 15.15 \\
Std Dev   &  2.61 & 2.95 & 6.78 & 6.60 & 4.35 & 3.93 & 5.52 & 3.06 & 3.71 \\
\bottomrule
\end{tabular}
\end{table}

\begin{figure}[h!b]
    \centering
    \begin{subfigure}[b]{0.49\textwidth}
        \centering
        \includegraphics[width=\textwidth]{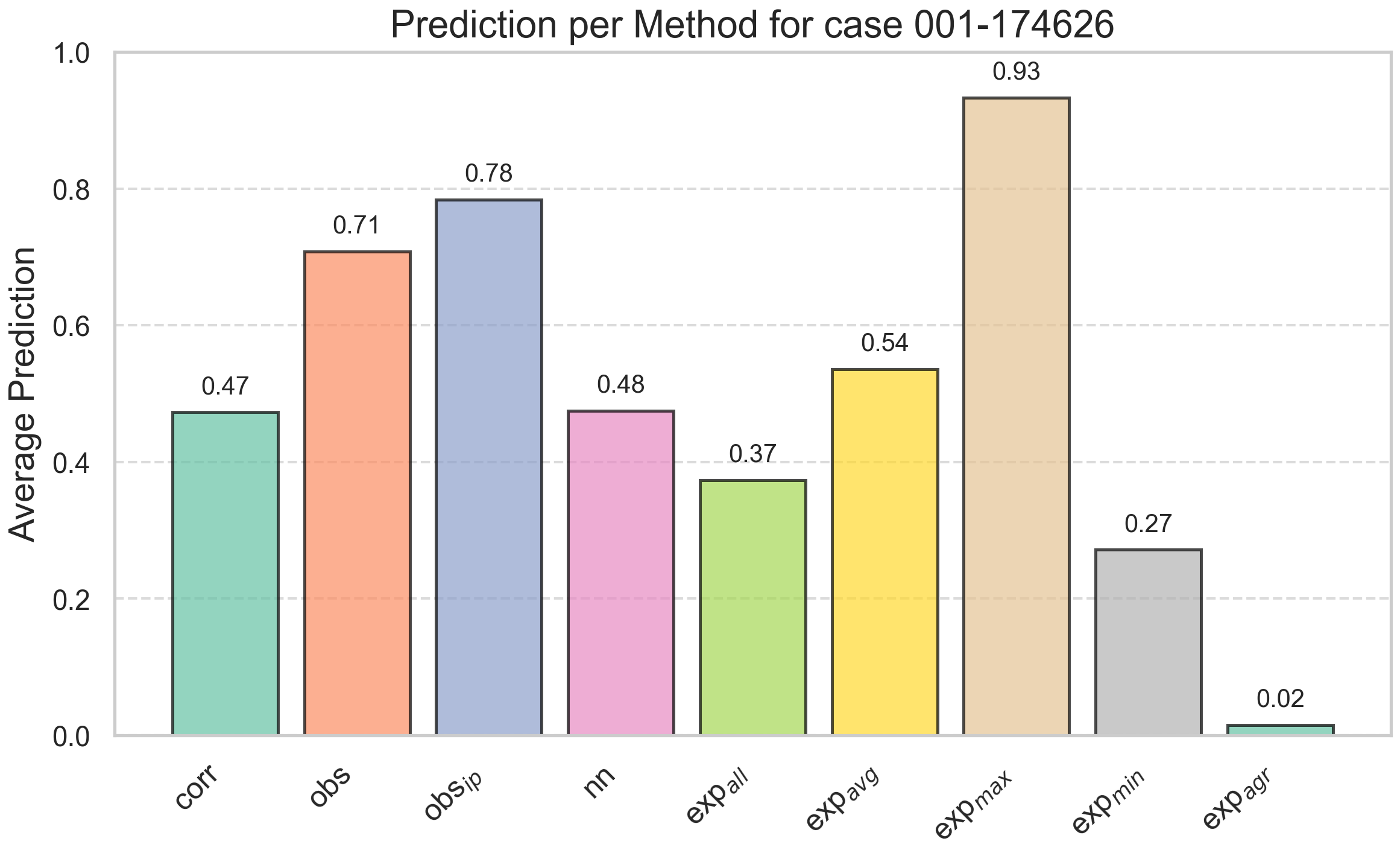}
        \caption{}
        \label{fig:comparison_1}
    \end{subfigure}
    \begin{subfigure}[b]{0.49\textwidth}
        \centering
        \includegraphics[width=\textwidth]{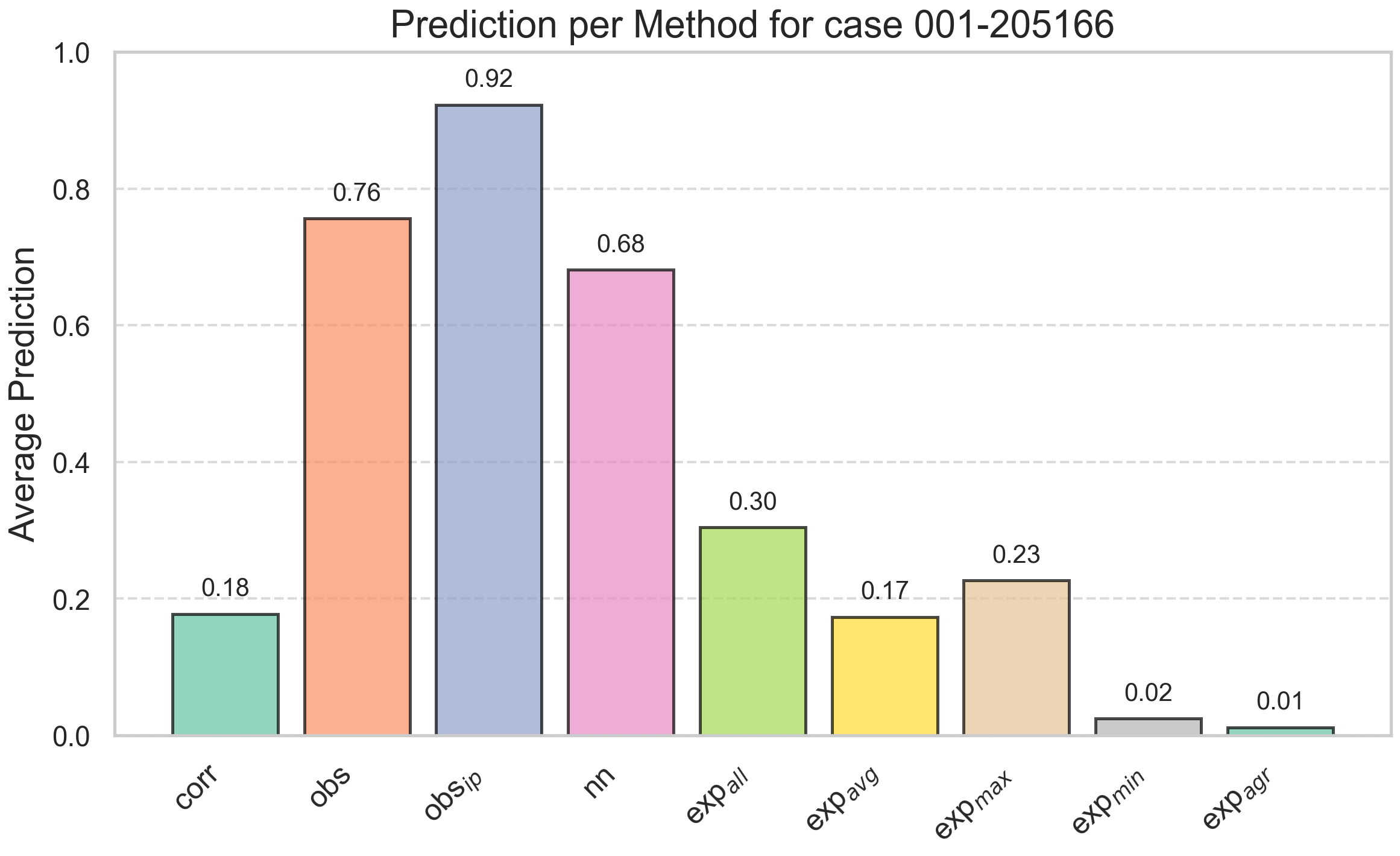}
        \caption{}
        \label{fig:comparison_2}
    \end{subfigure}
    \caption{The mean prediction for two specific cases using different label imputation methods.}
    \label{fig:individual_comparison}
\end{figure}

For each label imputation method, we average the results across the seven balanced train sets to investigate how it affects the behaviour of the model. Note again that we only consider Chamber Case decisions as ground truth.
Table~\ref{tbl:mcc_combined} reports the mean performance of the models on both test sets. In the top rows, we show the test set containing only Grand Chamber cases (34 cases). These serve as the ground truth and provide, to some degree, an objective measure of performance. 
The bottom rows of Table~\ref{tbl:mcc_combined} show results for 871 Chamber cases. Although performance is higher for this set, the outcomes of Chamber cases are considered indeterminate, so these figures cannot be interpreted as true performance measures. 
Figure~\ref{fig:violin} illustrates the distribution of predictions for models trained using different methods of imputing indeterminate labels on both test sets, showing that the choice of method influences the predictions of the model. 
In Figure~\ref{fig:individual_comparison}, we show the different predictions for two specific cases, further demonstrating the effects of the various label imputation methods.

\section{Discussion}
When interpreting the results, it is important to note that Chamber case outcomes are considered indeterminate, so we cannot determine whether the model’s predictions on these cases are correct. Moreover, all nine label imputation methods rely on unverifiable assumptions (Table~\ref{tbl:methods}), and none can be regarded as absolutely `correct'. We chose these methods from the existing literature primarily to illustrate the problem of label indeterminacy in law, though some of their underlying assumptions may be especially implausible in legal contexts. This paper does not aim to provide a best-performing method or model, but rather to show why ignoring label indeterminacy is misguided.

Table~\ref{tbl:mcc_combined} shows low MCC scores, indicating limited predictive performance. Prior work mostly reports accuracy and F1-scores~\cite{medvedeva-mcbride-2023-legal}, which do not fully capture all confusion matrix quadrants, a crucial consideration for test sets with heavily skewed label distributions (see Table~\ref{tbl:split_summary}). We also account for temporal effects by splitting the data into training and test sets based on judgment year, a setup shown to be more challenging~\cite{steging2023taking}. That study reported mean MCCs between 8.16 and 26.30 depending on the test year. Using the last nine years as our test set, we observe an average MCC of 14.43 on Chamber cases when training the model `as usual' ($corr$), which aligns with these expectations. Performance on Grand Chamber cases is generally lower and more variable, except for the $obs$ and $obs+ip$ methods, which use only Grand Chamber cases for training and achieve higher results on the Grand Chamber test set.

Figure~\ref{fig:violin} shows that the model’s predictive behaviour varies depending on the label imputation method used during training. For example, the $exp_{max}$ method generally produces high predictions, approaching 1.0, while the $nn$ method is more conservative, yielding lower predictive values. 

To further highlight the impact of label imputation on predictive behaviour, Figure~\ref{fig:individual_comparison} shows the average predictions of models trained with each of the nine methods for two example cases. In Figure~\ref{fig:comparison_1}, predictions range from 0.93 using the $exp_{max}$ method to 0.02 with $exp_{agr}$, a difference of 0.91, with the other methods falling in between. Similarly, Figure~\ref{fig:comparison_2} also shows a 0.91 difference, but between $obs+ip$ and $exp_{agr}$. Examining the two cases revealed no clear reason for the divergent predictions. Both are Chamber cases, one with a single dissent and the other decided by majority, suggesting that ordinary cases such as these can be strongly affected by how label indeterminacy is handled. 

Overall, our results highlight the importance of accounting for label indeterminacy, as the choice of method can substantially affect model behaviour. Predicting ECtHR case outcomes is challenging even for humans, as evidenced by dissenting opinions, and serves mainly as an academic illustration. 
We therefore view the experiment as an illustrative case study that shows how different label imputation strategies influence model predictions. 
Similar issues arise in other legal contexts, such as settlements, withdrawals, procedural defaults, bail, or appeals. 
We therefore argue that both future research and developers in the legal domain should explicitly consider label indeterminacy when designing machine learning models.

\section{Conclusion}
In this paper, we introduce the concept of label indeterminacy to the legal domain and highlight its critical implications for machine learning. Certain legal outcomes are inherently unknowable because they depend on interventions whose absence or alteration could have produced different results. This creates indeterminate labels that cannot be ignored in predictive modeling. Using the ECtHR as a case study, we show that commonly applied label imputation strategies can yield different prediction behaviours, emphasizing the need to explicitly address indeterminacy when designing and evaluating AI systems in law.

\section*{Acknowledgements}
This research was partially funded by the Hybrid Intelligence Center, a 10-year programme funded by the Dutch Ministry of Education, Culture and Science through the Netherlands Organisation for Scientific Research, https://hybrid-intelligence-centre.nl.  This research was also supported by Future Law Lab under the Strategic Programme Excellence Initiative at Jagiellonian University. The authors wish to express their gratitude to Jakob Schoeffer, Piotr Bystranowski, Maciej Próchnicki, Bartosz Janik, Michał Rachalski, and Alexander Stachurski for their valuable feedback and support during the preparation of this article.

\bibliographystyle{vancouver}
\bibliography{base}

\end{document}